\documentclass[sigconf]{acmart}
\usepackage{CJKutf8}
\usepackage{url}
\usepackage{booktabs}
\usepackage{enumitem}
\usepackage{makecell}
\usepackage{hyperref}
\usepackage{xcolor}
\usepackage{graphicx}
\usepackage[export]{adjustbox}
\usepackage{xspace}
\usepackage{multirow}
\usepackage{balance}

\usepackage{bbm}
\usepackage{amsfonts}

\usepackage[linesnumbered,ruled,vlined]{algorithm2e} %[ruled,vlined]{
\usepackage{algpseudocode}
\usepackage{amsmath}
\newtheorem{thm} {\bf Theorem}

\AtBeginDocument{%
  \providecommand\BibTeX{{%
    \normalfont B\kern-0.5em{\scshape i\kern-0.25em b}\kern-0.8em\TeX}}}

\copyrightyear{2022} 
\acmYear{2022} 
\setcopyright{acmcopyright}
\acmConference[CIKM '22] {Proceedings of the 31st ACM International Conference on Information and Knowledge Management}{October 17--21, 2022}{Atlanta, GA, USA.}
\acmBooktitle{Proceedings of the 31st ACM International Conference on Information and Knowledge Management (CIKM '22), October 17--21, 2022, Atlanta, GA, USA}
\acmPrice{15.00}
\acmISBN{978-1-4503-9236-5/22/10}
\acmDOI{10.1145/3511808.3557626}
\begin{document}
\title{Learning Rate Perturbation: A Generic Plugin of Learning Rate schedule towards Flatter Local Minima}

\author{Hengyu Liu}
\authornote{Work done while this author was an intern at Microsoft Research}
\email{hengyuliu94@gmail.com}
\orcid{1234-5678-9012}
\affiliation{%
  \institution{Northeastern University}
  \city{Shenyang}
  \state{Liaoning}
  \country{China}
}

\author{Qiang Fu}
\authornote{Corresponding Author}
\email{qifu@microsoft.com}
\affiliation{%
  \institution{Microsoft Research Asia}
  \city{Beijing}
  \country{China}}

\author{Lun Du}
\email{lun.du@microsoft.com}
\affiliation{%
  \institution{Microsoft Research Asia}
  \city{Beijing}
  \country{China}}

\author{Tiancheng Zhang}
\email{tczhang@mail.neu.edu.cn}
\affiliation{%
  \institution{Northeastern University}
  \city{Shenyang}
  \state{Liaoning}
  \country{China}
}

\author{Ge Yu}
\email{yuge@mail.neu.edu.cn}
\affiliation{%
  \institution{Northeastern University}
  \city{Shenyang}
  \state{Liaoning}
  \country{China}
}

\author{Shi Han}

\email{shihan@microsoft.com}
\affiliation{%
  \institution{Microsoft Research Asia}
  \city{Beijing}
  \country{China}}
  
\author{Dongmei Zhang}
\email{dongmeiz@microsoft.com}
\affiliation{%
  \institution{Microsoft Research Asia}
  \city{Beijing}
  \country{China}}

\vspace{-5mm}


\begin{abstract}
Learning rate is one of the most important hyper-parameters that has significant influence for neural network training. 
Learning rate schedules are widely used in real practice to adjust the learning rate according to pre-defined schedules for the fast convergence and good generalization.
However, existing learning rate schedules are all heuristic algorithms and lack theoretical support. 
Therefore, people usually choose the learning rate schedules through multiple ad-hoc trial, and the obtained learning rate schedules are sub-optimal. 
To boost the performance of the obtained sub-optimal learning rate schedule, we propose a generic learning rate schedule plugin, called \textbf{LEA}rning Rate \textbf{P}erturbation (LEAP), which can be applied to various learning rate schedules to improve the model training by introducing a certain perturbation to the learning rate.
We found that, with such simple yet effective strategy, training processing exponentially favors flat minima rather than sharp minima with guaranteed convergence, which leads to better generalization ability.  
In addition, we conduct extensive experiments which show that training with LEAP can improve the performance of various
deep learning models on diverse datasets using various learning rate schedules (including constant learning rate). 
\end{abstract}






\maketitle
\vspace{-4mm}
\section{Introduction}
\label{sec:org4deb1a6}
Deep neural networks are the basis of state-of-the-art results for broad tasks such as image recognition, speech recognition, machine translation, driver-less car technology, source code understanding, social analysis, and tabular data understanding \cite{mao2021neuron,fu2021neuron,shi2021neural,shi2021cast,song2020inferring,du2021tabularnet}. It is known that the learning rate is one of the most important hyper-parameters which has significant influence for training deep neural networks.
Too big learning rate may result in the great difficulty of finding minima or even the non-convergence issue, while too small learning rate greatly slows down the training process and may easily get stuck in sharp minima.
How to adjust the learning rate is one of the challenges of training deep learning models.

Learning rate schedules seek to adjust the learning rate during training process by changing the learning rate according to a pre-defined schedule. Various learning rate schedules have been widely used and shown the practical effectiveness for better training. The existing learning rate schedules can be roughly divided into several categories, constant learning rate schedule, learning rate decay schedule \cite{you2019does}, learning rate warm restart schedule \cite{smith2017cyclical, loshchilov2016sgdr,mishra2019polynomial} and adaptive learning rate schedule \cite{preechakul2019cprop,zeiler2012adadelta,liu2019variance,lewkowycz2021decay}. 
However, the existing learning rate schedules are all heuristic algorithms and lack theoretical support. It is because that the mapping functions of DNN models are usually with high complexity and non-linearity, and it is quite difficult to analyze the influence of the learning rate on the training process. Therefore, people may choose a learning rate schedule only through multiple ad-hoc trial, and the obtained learning rate schedule are probably sub-optimal.

As an effective approach to improve generalization ability of DNN models, perturbation addition has been widely studied \cite{reed1999neural}.
It has been demonstrated that training with noise can indeed lead to improvements in network generalization \cite{bishop1995neural}.
Existing works attempt to introduce perturbation in the feature space \cite{bishop1995training}, activation function  \cite{gulcehre2016noisy}, and gradient \cite{neelakantan2015adding}.
\cite{bishop1995training} shows that adding noise to feature space during training is equivalent to Tikhonov regularization.
\cite{gulcehre2016noisy} proposes to exploit the injection of appropriate perturbation to the layer activations so that the gradients may flow easily (i.e., mitigating the vanishing gradient problem).
\cite{neelakantan2015adding} adds perturbation to gradients to improve the robustness of the training process. The amount of noise can start high at the beginning of training and decrease over time, much like a decaying learning rate. 
However, there is no existing work that attempts to introduce randomness into hyper-parameters (e.g. learning rate).

To boost the obtained sub-optimal learning rate schedule, we propose a generic plugin of learning rate schedule, called Learning Rate Perturbation (LEAP), which can be applied to various learning rate schedules to improve the model training by introducing a certain type of perturbation to the learning rate.
We leverage existing theoretical framework \cite{xie2020diffusion} to analyze the impact of LEAP on the training process, and found that training process applied with LEAP favors flatter minima exponentially more than sharpness minima. It is well studied that learning flat minima closely relate to generalization.

In deep learning models, there are already some generic strategies or plugins that can boost training performance, such as Dropout, Batch Normalization. The reason why we consider them as a kind of plugins is they could be applied to a particular aspect of DNN learning with a wide range of versatility. In more details, Dropout could be considered as a generic plugin applied to the network structure of deep learning models, which prevents deep learning models from over-reliance on a single neural unit by randomly ignoring some weights during training processing. Batch Normalization is a generic plugin applied to layer input/output, which improves the performance of the model by normalizing the value range of intermediate layer input/output. With different perspective from Dropout and Batch Normalization, LEAP is a generic plugin applied to the learning rate, which improves model performance by letting the training process favors flatter minima. 

The main contributions of this work are outlined as follows:
\begin{itemize}
    \item We propose a simple yet effective plugin of learning rate schedule, called \textbf{LEA}rning Rate \textbf{P}erturbation (LEAP), which can be applied to various learning rate schedules to improve the model training by introducing a certain perturbation to the learning rate. 
    
    \item To the best of our knowledge, our work is the first one to propose a generic strategy with theoretical guarantees to boost training performance of various learning rate schedule by letting training process favor flat minima.
    
    \item The extensive experiments show that LEAP can effectively improve the training performance of various DNN architectures, including Multi-Layer Perceptron (MLP), Convolutional Neural Network (CNN), Graph Neural Network (GNN), and Transformer, with different learning rate schedules and optimizers on diverse domains.
\end{itemize}
\vspace{-2mm}
\section{Learning Rate Perturbation}
\label{sec:org7c9e4fa}

In this section, we introduce the implementation details of LEAP, and given the pseudo code of training process applied with LEAP.

Firstly, we denote the training dataset as \(x\), one batch of training datasets as \(x_{j}\), the learning rate given by the learning rate schedule as \(\eta\), the model parameters as \(\theta\),  and the loss function as \(L(\theta,x)\). For simplicity, we denote the training loss as \(L(\theta)\).
Our LEAP is to add perturbation satisfying Gaussian distribution, i.e., $\zeta \sim \mathcal{N}(0, \sigma^{2})$, to the learning rate \(\eta\). Here, \(\sigma\) is the hyper-parameter of LEAP to control perturbation intensity.
The training process applied with LEAP for updating parameters is shown in Algorithm \ref{algorithm:LEAP-SGD}.

Here, we define the learning rate after applying LEAP as a vector  \(\mathbf{h} = (\eta_1, ..., \eta_M)\),  $\eta_i$ is the learning rate used for updating \textit{i}-th parameter, which is calculated by adding the learning rate perturbation $\zeta_i$ to the original learning rate \(\eta\) given by the learning rate schedule, and \(|\mathbf{h}| = |\theta| = M\) (\(M\) is the total number of parameters).
Note that, the perturbations on different feature dimensions ($\zeta_1,\zeta_2,...,\zeta_M$) are independent so that the overall perturbations can lie along with diverse directions. 
Obviously, the learning rate \(\mathbf{h}\) in LEAP obeys a high-dimensional Gaussian distribution in the following form:
\begin{equation}
\label{eq:1}
\mathbf{h} \sim \mathcal{N}(\eta \mathbf{i}, \eta^{2} \sigma^{2} \mathbf{I})
\end{equation}
where \(\mathbf{i}\) is a vector where all elements are 1, \(\mathbf{I}\) is the identity matrix.

By combining Eq. \ref{eq:2}, the parameter update formula of training process applied with LEAP is as follows.
\begin{equation}
\begin{aligned}
\label{eq:2}
\theta_{t+1} &= \theta_{t} - \mathbf{h} \circ \frac{\partial L(\theta_{t},x_{j})}{\partial \theta_{t}}
= \theta_t - \eta \frac{\partial L(\theta_{t},x_{j})}{\partial \theta_{t}} + \eta A(\theta) \zeta_{t} \\
\end{aligned}
\end{equation}
\begin{equation}
\label{eq:4}
A(\theta) = diag(\frac{\partial L \left( \theta_t, x_{j} \right)}{\partial \theta_{t, 1}}, ..., \frac{\partial L \left( \theta_t, x_{j} \right)}{\partial \theta_{t, M}})
\end{equation}
where \(L(\theta, x_{j})\) is the loss of the \textit{j}-th mini-batch, \(\circ\) is the Hadamard product, \(\frac{\partial L(\theta_{t},x_{j})}{\partial \theta_{t, i}}\) is the gradient of the \textit{i}-th parameter in the \textit{t}-th update, and \(\zeta_{t} \sim \mathcal{N}(0, \sigma^2 \textbf{I})\) which is a zero-mean high dimensional Gaussian distribution.

\begin{algorithm}[!htbp]
\caption{The training process applied with LEAP}
\label{algorithm:LEAP-SGD}
\KwIn{ training dataset $x$;  \\ \qquad \quad learning rate schedule function $LRS$; \\ \qquad \quad init parameters $\theta$}
\KwOut{trained parameters $\theta$}
\For {e = 1 to IterNum}
{
    $\eta_e = LRS(e)$
    \For {j = 1 to $B$}
   {
        Get a batch of training data $x_{j}$ from training dataset \\
        Sample $\mathbf{h} \sim \mathcal{N}(\eta_e \mathbf{i}, \eta^{2} \sigma^{2} \mathbf{I})$ \\
        Update $\theta = \theta - \mathbf{h} \circ \frac{\partial L(\theta,x_{j})}{\partial \theta}$
   }
}
\end{algorithm}
%\vspace{-8mm}

\begin{table*}[!ht]
    \centering
    \begin{tabular}{c | l|ccc|ccc}
    \hline
    & Optimizer              &      \multicolumn{3}{c|}{SGD}  &  \multicolumn{3}{c}{Adam}  \\
    \hline
    &                        & Vanilla &  LEAP(Ours) & \textbf{Gain} & Vanilla & LEAP (Ours) & \textbf{Gain} \\
    \hline
    \multirow{3}{*}{ \centering \rotatebox{90}{\centering  Cifar-10}}
    % Cifar-10                &   &&&&&\\
    &\quad  ResNet-18        &  5.60 $\pm$ 0.11  & 5.32 $\pm$ 0.25  & \quad  \textbf{5.00\%} \quad      
                            &  9.26 $\pm$ 0.29  & 8.69 $\pm$ 0.08  & \quad  \textbf{6.16\%} \quad  \\
    &\quad  ResNet-50        &  5.26 $\pm$ 0.07  & 4.78 $\pm$ 0.13  & \quad  \textbf{9.13\%} \quad     
                            &  7.84 $\pm$ 0.52  & 7.17 $\pm$ 0.12  & \quad  \textbf{8.78\%} \quad  \\
    &\quad  ResNet-101       &  5.09 $\pm$ 0.09  & 4.69 $\pm$ 0.08  & \quad  \textbf{7.86\%} \quad      
                            &  7.13 $\pm$ 0.75  & 6.62 $\pm$ 0.04  & \quad  \textbf{7.15\%} \quad   \\
    \hline
    \end{tabular}
    \caption{Error rate (\%) of applying our LEAP to 3 CNN models on Cifar-10 dataset with two optimizers (lower indicators are better). The columns leaded by the cell ``Gain'' present relative improvements.}
    \label{tab:Optimizer}
\end{table*}

\begin{table}[!ht]
    \centering
    % \small
    \begin{tabular}{c|l|ccc}
    \hline
    &                        & Vanilla &  LEAP(Ours) & \textbf{Gain} \\
    \hline
    \multirow{5}{*}{ \centering \rotatebox{90}{\centering  MNIST}}
    % MNIST                   &   &&\\
    &\quad  MLP-3            &  2.24 $\pm$ 0.09  &  2.02 $\pm$ 0.07  &   \textbf{9.82\%}  \\
    &\quad  MLP-4            &  2.23 $\pm$ 0.06  &  2.04 $\pm$ 0.04  &   \textbf{8.52\%}  \\
    &\quad  MLP-6            &  2.24 $\pm$ 0.05  &  1.99 $\pm$ 0.13  &   \textbf{11.16\%}  \\
    &\quad  MLP-8            &  2.54 $\pm$ 0.09  &  2.26 $\pm$ 0.10  &   \textbf{11.02\%}   \\
    &\quad  MLP-10            &  2.44 $\pm$ 0.10  &  2.25 $\pm$ 0.9  &  \textbf{7.79\%}   \\
    \hline
    \multirow{3}{*}{ \centering \rotatebox{90}{\centering  Cifar-10}}
    %Cifar-10                &   &&\\
    &\quad  ResNet-18        &  7.11 $\pm$ 0.17  & 6.78 $\pm$ 0.06 &   \textbf{4.64\%}  \\
    &\quad  ResNet-50        &  6.93 $\pm$ 0.10  & 6.57 $\pm$ 0.18 &   \textbf{5.19\%}  \\
    &\quad  ResNet-101       &  6.67 $\pm$ 0.08  & 6.26 $\pm$ 0.20 &   \textbf{6.15\%}  \\
    \hline
    \multirow{2}{*}{ \centering \rotatebox{90}{\centering  IN}}
    % ImageNet                &   &&\\
    % &&&\\ 
    &\quad  ResNet-50        &  28.38 $\pm$ 0.09 &  26.74 $\pm$ 0.15  &  \textbf{5.78\%} \\
    &\quad  VGG-19           &  31.17 $\pm$ 0.11 &  29.51 $\pm$ 0.17  &  \textbf{5.33\%} \\
    % &&&\\
    \hline
    \end{tabular}
    \caption{Error rate (\%) of applying our LEAP to 5 MLP models, and 4 CNN models on three CV datasets without learning rate schedule (lower indicators are better). The columns leaded by the cell ``Gain'' present relative improvements.}
    \label{tab:CVDomainWithoutLRS}
\end{table}

\begin{table}[!ht]
    \centering
    \begin{tabular}{c|l|ccc}
    \hline
    &                 & Vanilla & LEAP (Ours) & \quad \textbf{Gain} \quad \\
    \hline
    \multirow{4}{*}{ \centering \rotatebox{90}{\centering  Cora}}
    % Cora                    &   &&\\
    &\quad  GCN       & 25.38 $\pm$  0.75 & 22.60 $\pm$  0.57 &  \textbf{10.95\%}  \\
    &\quad  GAT       & 29.72 $\pm$  1.21 & 26.18 $\pm$  1.71 &  \textbf{11.91\%}  \\
    &\quad  GIN       & 37.46 $\pm$  2.93 & 33.02 $\pm$  2.71 &  \textbf{11.85\%}  \\
    &\quad  GraphSage & 26.24 $\pm$  0.45 & 23.50 $\pm$  0.66 &  \textbf{10.44\%}  \\
    \hline
    \multirow{4}{*}{ \centering \rotatebox{90}{\centering  PubMed}}
    % PubMed                  &   &&\\
    &\quad  GCN       & 25.96 $\pm$  0.73  & 23.98 $\pm$  1.00 &  \textbf{7.63\%}  \\
    &\quad  GAT       & 26.60 $\pm$  0.49  & 24.90 $\pm$  1.63 &  \textbf{6.39\%}  \\
    &\quad  GIN       & 29.24 $\pm$  1.33  & 27.22 $\pm$  1.87 &  \textbf{6.91\%}  \\
    &\quad  GraphSage & 27.02 $\pm$  0.33  & 25.26 $\pm$  1.26 &  \textbf{6.51\%}  \\
    \hline
    \multirow{4}{*}{ \centering \rotatebox{90}{\centering  CiteSeer}}
    % CiteSeer                &   &&\\
    &\quad  GCN        & 37.38 $\pm$  1.36  & 33.68 $\pm$  0.96 &  \textbf{9.90\%}  \\
    &\quad  GAT        & 37.96 $\pm$  3.65  & 35.36 $\pm$  2.01 &  \textbf{6.85\%}  \\
    &\quad  GIN        & 49.38 $\pm$  3.30  & 46.02 $\pm$  2.18 &  \textbf{6.80\%}  \\
    &\quad  GraphSage  & 38.02 $\pm$  0.90  & 34.64 $\pm$  0.38 &  \textbf{8.89\%}  \\
    \hline
    \end{tabular}
    \caption{Error rate (\%) of applying our LEAP to 4 GNN models on three Graph datasets without learning rate schedule (lower indicators are better). The columns leaded by the cell ``Gain'' present relative improvements.}
    \label{tab:GNNDomainWithoutLRS}
\end{table}

\begin{table*}[!ht]
    \centering
    \begin{tabular}{c|l|ccc|ccc}
    \hline
    &\, Learning Rate Schedule  &      \multicolumn{3}{c|}{Decay}  &  \multicolumn{3}{c}{Warm Restart}  \\
    \hline
    &                        & Vanilla &  LEAP(Ours) & \textbf{Gain} & Vanilla & LEAP (Ours) & \textbf{Gain}     \\
    \hline
    \multirow{5}{*}{ \centering \rotatebox{90}{\centering  MNIST}}
    % MNIST                   &   &&&&&\\
    &\quad  MLP-3            &  2.48 $\pm$ 0.06  &  2.22 $\pm$ 0.04  &  \quad \textbf{10.48\%} \quad  
                            &  2.34 $\pm$ 0.05  &  2.12 $\pm$ 0.03  &  \quad \textbf{9.40\%} \quad  \\
    &\quad  MLP-4            &  1.96 $\pm$ 0.09  &  1.79 $\pm$ 0.06  &  \quad \textbf{8.67\%} \quad  
                            &  2.20 $\pm$ 0.11  &  2.02 $\pm$ 0.11  &  \quad \textbf{8.18\%} \quad  \\
    &\quad  MLP-6            &  2.24 $\pm$ 0.04  &  2.01 $\pm$ 0.13  &  \quad  \textbf{10.27\%} \quad  
                            &  2.20 $\pm$ 0.20  &  1.99 $\pm$ 0.04  &  \quad \textbf{9.55\%} \quad  \\
    &\quad  MLP-8            &  2.54 $\pm$ 0.09  &  2.25 $\pm$ 0.10  &  \quad  \textbf{11.42\%} \quad                            &  2.62 $\pm$ 0.07  &  2.41 $\pm$ 0.08  &  \quad  \textbf{8.02\%} \quad \\
    &\quad  MLP-10           &  2.49 $\pm$ 0.12  &  2.15 $\pm$ 0.10  &  \quad  \textbf{13.65\%} \quad                            &  2.25 $\pm$ 0.05  &  1.96 $\pm$ 0.09  &  \quad  \textbf{12.89\%} \quad \\
    \hline
    \multirow{3}{*}{ \centering \rotatebox{90}{\centering  Cifar-10}}
    %Cifar-10                &   &&&&&\\
    &\quad  ResNet-18        &  5.60 $\pm$ 0.11  & 5.32 $\pm$ 0.25  & \quad  \textbf{5.00\%} \quad      
                            &  5.73 $\pm$ 0.09  & 5.27 $\pm$ 0.10  & \quad  \textbf{8.03\%} \quad  \\
    &\quad  ResNet-50        &  5.26 $\pm$ 0.07  & 4.78 $\pm$ 0.13  & \quad  \textbf{9.13\%} \quad     
                            &  5.38 $\pm$ 0.16 &  4.86 $\pm$ 0.06  & \quad  \textbf{9.67\%} \quad  \\
    &\quad  ResNet-101       &  5.09 $\pm$ 0.09  & 4.69 $\pm$ 0.08  & \quad  \textbf{7.86\%} \quad      
                            &  5.19 $\pm$ 0.09 &  4.68 $\pm$ 0.14  & \quad  \textbf{9.83\%} \quad   \\
    \hline
    \multirow{3}{*}{ \centering \rotatebox{90}{\centering  IN}}
    %ImageNet                &   &&&&&\\
    % &&&&&&&\\
    &\quad  ResNet-50        &  24.48 $\pm$ 0.07 &  22.87 $\pm$ 0.04  & \quad \textbf{6.58\%} \quad 
                            &  25.24 $\pm$ 0.21 &  23.51 $\pm$ 0.12  & \quad \textbf{6.85\%} \\
    &\quad  VGG-19           &  28.87 $\pm$ 0.10 &  27.33 $\pm$ 0.09  & \quad \textbf{5.33\%} \quad 
                            &  29.17 $\pm$ 0.16 &  27.47 $\pm$ 0.08  & \quad \textbf{5.83\%} \\
    &\quad  Swin Transformer &       -           &          -         &         -                   
                            &  18.86 $\pm$ 0.05 &  17.86 $\pm$ 0.03  & \quad \textbf{5.30\%} \quad \\
    \hline
    \end{tabular}
    \caption{Error rate (\%) of applying our LEAP to 5 MLP models, 4 CNN models and 1 Transformer model on three CV datasets with two learning rate schedule (lower indicators are better). The columns leaded by the cell ``Gain'' present relative improvements.}
    \label{tab:CVDomain}
\end{table*}

\begin{table*}[!ht]
    \centering
    \begin{tabular}{c|l|ccc|ccc}
    \hline
    &  Learning Rate Schedule  &      \multicolumn{3}{c|}{Decay}  &  \multicolumn{3}{c}{Warm Restart}  \\
    \hline
    &                 & Vanilla & LEAP (Ours) & \quad \textbf{Gain} \quad & Vanilla & LEAP (Ours) & \quad \textbf{Gain} \quad     \\
    \hline
    \multirow{4}{*}{ \centering \rotatebox{90}{\centering  Cora}}
    % Cora                    &   &&&&&\\
    &\quad  GCN       & 25.84 $\pm$  0.84 & 23.18 $\pm$  1.40 & \quad \textbf{10.29\%} \quad 
                     & 25.50 $\pm$  0.71 & 22.88 $\pm$  0.72 & \quad \textbf{10.27\%} \quad \\
    &\quad  GAT       & 28.80 $\pm$  1.10 & 25.98 $\pm$  2.08 & \quad \textbf{9.79\%} \quad 
                     & 29.76 $\pm$  1.25 & 26.68 $\pm$  1.72 & \quad \textbf{10.35\%} \quad \\
    &\quad  GIN       & 38.06 $\pm$  3.44 & 33.58 $\pm$  1.71 & \quad \textbf{11.77\%} \quad 
                     & 37.18 $\pm$  1.48 & 34.38 $\pm$  2.06 & \quad \textbf{7.53\%} \quad \\
    &\quad  GraphSage & 26.64 $\pm$  0.35 & 23.66 $\pm$  0.42 & \quad \textbf{11.19\%} \quad 
                     & 27.78 $\pm$  0.88 & 23.54 $\pm$  0.69 & \quad \textbf{15.26\%} \quad \\
    \hline
    \multirow{4}{*}{ \centering \rotatebox{90}{\centering  PubMed}}
    % PubMed                  &   &&&&&\\
    &\quad  GCN       & 28.40 $\pm$  0.97  & 25.20 $\pm$  0.70 & \quad \textbf{11.27\%} \quad 
                     & 27.24 $\pm$  1.33 & 25.18 $\pm$  0.32  & \quad \textbf{7.56\%} \quad \\
    &\quad  GAT       & 27.86 $\pm$  0.71  & 26.16 $\pm$  0.63 & \quad \textbf{6.10\%} \quad 
                     & 28.88 $\pm$  1.62 & 26.96 $\pm$  1.02  & \quad \textbf{6.65\%} \quad \\
    &\quad  GIN       & 29.54 $\pm$  1.77  & 27.46 $\pm$  1.00 & \quad \textbf{7.04\%} \quad 
                     & 31.36 $\pm$  1.33 & 28.86 $\pm$  2.43  & \quad \textbf{7.97\%} \quad \\
    &\quad  GraphSage & 29.46 $\pm$  0.97  & 25.98 $\pm$  0.49 & \quad \textbf{11.81\%} \quad 
                     & 28.18 $\pm$  1.07 & 25.30 $\pm$  0.72 & \quad \textbf{10.22\%} \quad\\
    \hline
    \multirow{4}{*}{ \centering \rotatebox{90}{\centering  CiteSeer}}
    % CiteSeer                &   &&&&&\\
    &\quad  GCN        & 36.70 $\pm$  0.73  & 33.90 $\pm$  1.13 & \quad \textbf{7.63\%} \quad 
                      & 37.96 $\pm$  0.85 & 34.24 $\pm$  1.00 & \quad \textbf{9.80\%} \quad \\
    &\quad  GAT        & 38.46 $\pm$  4.04  & 34.72 $\pm$  2.04 & \quad \textbf{9.72\%} \quad 
                      & 37.92 $\pm$  4.41 & 34.94 $\pm$  1.42 & \quad \textbf{7.86\%} \quad \\
    &\quad  GIN        & 53.02 $\pm$  4.57  & 48.96 $\pm$  2.42 & \quad \textbf{7.66\%} \quad 
                      & 50.88 $\pm$  4.86 & 45.10 $\pm$  2.97 & \quad \textbf{11.36\%} \quad \\
    &\quad  GraphSage  & 39.82 $\pm$  1.47  & 35.68 $\pm$  0.61 & \quad \textbf{10.40\%} \quad 
                      & 38.72 $\pm$  0.45 & 34.58 $\pm$  1.09 & \quad \textbf{10.69\%} \quad \\
    \hline
    \end{tabular}
    \caption{Error rate (\%) of applying our LEAP to 4 GNN models on three Graph datasets with two learning rate schedules (lower indicators are better). The columns leaded by the cell ``Gain'' present relative improvements.}
    \label{tab:GNNDomain}
\end{table*}
\section{Minima Preference Analysis of LEAP}
By exploiting \cite{xie2020diffusion} theoretical framework, we replace stochastic gradient noise with $\eta A(\theta) \zeta_{t}$ and obtain Theorem \ref{EscapesMinima}.
 Theorem \ref{EscapesMinima} present analysis result on escape time of LEAP at any minima \( a\).

\begin{thm}
\label{EscapesMinima}
LEAP Escape Time at Minima. The loss function $L(\theta)$ is of class $C^2$ and N-dimensional. 
If the dynamics is governed by equation \ref{eq:2}, then the mean escape time from minima $a$ to the outside of minima $a$ is

\begin{equation}
\notag
t =2 \pi C \frac{1}{\left|H_{b e}\right|} \exp \left[\frac{2 \Delta L}{\eta \sigma^2 }\left(\frac{s}{A_{a e}}+\frac{(1-s)}{\left|A_{b e}\right|}\right)\right]
\end{equation}

where $C = \sqrt{\frac{det(H_a (diag(H_a))^{-1})}{-det(H_b (diag(H_b))^{-1})}}$, $s \in (0, 1)$ is a path-dependent parameter, $H_{be}$ is the eigenvalues of Hessian matrix $H(\theta)$ at the saddle point $b$ corresponding to the escape direction $e$, and $A_{ae}$ and $A_{be}$ are, respectively, the eigenvalues of $diag(H(\theta))$ at the minima $a$ and the saddle point $b$ corresponding to the escape direction $e$.
\end{thm}

\noindent \textbf{Explanation.} We can see that the mean escape time exponentially depends on \(A_{ae}\) which is the eigenvalue of $diag(H(\theta))$ at minima along the escape direction.
The flatter the minima, the smaller the eigenvalue of $diag(H(\theta))$, and the longer the training process stays in the minima.
Therefore, we conclude that LEAP stays in flatter minima exponentially longer than the sharp minima.

\noindent \textbf{Minima selection.} By exploiting \cite{xie2020diffusion} theoretical framework, we can formulate the probability of converged minima as $P(\theta \in V_a) = \frac{t_a}{\sum_v t_v}$. 
In deep learning, the landscape contain many good minima and bad minima. Training process transits from one minima to another minima. 
The mean escape time of one minima corresponds to the number of updates which training process spends on this minima.
Therefore, escape time is naturally proportional to the probability of selecting this minima.  
So we conclude that LEAP favors flat minima exponentially more than sharp minima.

\noindent \textbf{Convergence.} 
Since the expectation of perturbation in LEAP is 0, i.e. $E(\eta A(\theta) \zeta_{t}) = 0$, LEAP can guarantees that $E(||\theta_{t+1} - \theta^*||) \leq ||\theta_{t} - \theta^*||$ decreases with $\eta$ when convex function $L$ is $\beta$ smooth and $\eta < \frac{1}{\beta}$  ($\eta$ is learning rate, $\theta_t$ is the current point and $\theta^*$ is the global minima). 
When the perturbation is not too large, LEAP does not affect the convergence. 
During training, we can ensure convergence by choosing appropriate $\sigma$.

\section{Experiment}

\subsection{Experiment Setup}
\noindent \textbf{Datasets.} We take 6 real datasets from Computer Vision and Graph Learning domain to evaluate our method, and adopt Cora, CiteSeer and PubMed \cite{yang2016revisiting} for Graph Learning domain \cite{du2022understanding,chen2021fast,du2018traffic}, MNIST \cite{lecun1998mnist}, CIFAR-10 \cite{krizhevsky2009learning} and ImageNet \cite{deng2009imagenet} (Denoted as IN in the experiment) for Computer Vision domain. 
Cora, CiteSeer and PubMed \cite{yang2016revisiting} are citation networks based datasets. 
MNIST is one of the most researched datasets in machine learning, and is used to classify handwritten digits. 
The CIFAR-10 and ImageNet dataset is a collection of images that are commonly used to train machine learning and computer vision algorithms.
For graph datasets, we use the public data splits provided by \cite{yang2016revisiting}.
For the MNIST dataset, we keep the test set unchanged, and we randomly select 50,000 training images for training and the other 10,000 images as validation set for hyper-parameter tuning.
For Cifar-10 and ImageNet dataset, we use the public data splits provided by \cite{krizhevsky2009learning,russakovsky2015imagenet}.

\noindent \textbf{Network Architechture.} To verify the effectiveness of our method across different neural architectures, we employ Multilayer Perceptrons (MLPs), Convolutional Neural Networks (CNNs), Transforms, and Graph Neural Networks (GNNs) \cite{du2022gbk} for evaluation. And we adopt ResNet-18,50,101 \cite{he2016deep}, VGG-19 \cite{simonyan2014very} for CNN, GCN \cite{kipf2016semi}, GAT \cite{velivckovic2017graph}, GIN \cite{xu2018powerful}, and GraphSage \cite{DBLP:journals/corr/HamiltonYL17} for GNN, Swin Transfermer \cite{liu2021swin} for Transforms.
We run five MLP models which are denoted as MLP-L, L indicates the number of layers including input layer.

\noindent \textbf{Learning Rate Schedules and Optimizers.} We verify that LEAP on three different learning rate schedules (Constant Learning Rate, Learning Rate Decay and Warm Restart Schedule \cite{loshchilov2016sgdr}).
In addition, we also explored the performance improvement of LEAP for two optimizers (SGD \cite{ruder2016overview} and Adam \cite{kingma2014adam}).

\noindent \textbf{Hyperparameters Settings.} For the models without classical hyperparameter settings, we performed a hyperparameter search to find the best one for the baseline model. 
And the hyperparameter search process ensure each model in the same domain has the same search space. 
For all the GNN models, learning rate is searched in $\{$0.1, 0.05, 0.01, 5e-3, 1e-3$\}$, and weight decay is searched in $\{$0.05, 0.01, 5e-3, 1e-3, 5e-4, 1e-4$\}$.
For ResNet and MLP, learning rate is searched in $\{0.1, 0.05, 0.01\}$, and weight decay is searched in $\{$5e-3, 1e-3, 5e-4, 1e-4$\}$.
Above is the hyperparameter setting and hyperparameter search for baselines. 
For LEAP, we search for the hyperparameter $\sigma$ in $\{$0.1, 0.05, 0.01, 5e-3, 1e-3, 5e-4, 1e-4$\}$ while keeping other hyperparameters unchanged. 
Note that we do not apply constant learning rate and learning rate decay for Swim Transformer because that these two learning rate schedules are not suitable for Swim Transformer. 
We run each experiment on five random seeds and make an average. Model training is done on Nvidia Tesla V100 GPU.

\vspace{-5mm}
\subsection{Experiment Results}
We evaluate the effectiveness of our methods from four dimensions: 
(1) different learning rate schedules; 
(2) different neural architectures; 
(3) datasets from different domains; 
(4) different optimizers.

Under the setting without using learning rate schedule, i.e., equally to constant learning rate schedule, the error rates of the image classification task for the vanilla models with and without LEAP are shown in Table~\ref{tab:CVDomainWithoutLRS}, and graph node classification task result is in Table~\ref{tab:GNNDomainWithoutLRS}.
Under the setting with using learning rate schedule i.e., learning rate decay and warm restart schedule, the error rates of the image classification task are shown in Table~\ref{tab:CVDomain}, and graph node classification task result is in Table~\ref{tab:GNNDomain}.
Table~\ref{tab:Optimizer} show that the error rates of the image classification task under different optimizers.
Combining the results in Table~\ref{tab:CVDomainWithoutLRS}, \ref{tab:GNNDomainWithoutLRS}, \ref{tab:CVDomain}, \ref{tab:GNNDomain} and \ref{tab:Optimizer}, we can conclude that LEAP brings consistent improvements in these four dimensions. 
First, LEAP is a generic plugin, which can work with all learning rate schedules including Learning Rate Decay, and Warm Restart Schedule. Besides, we can see that LEAP can improve the performance of deep learning models even without basic learning rate schedules.
We obtain 5.00 \% to 13.65\% relative error rate reduction with learning rate decay, and 6.65 \% to 15.26\% with Warm Restart Schedule. 
Second, LEAP brings consistent improvements across all neural architectures including MLPs, CNNs, Transformer, and GNNs. 
Third, vanilla models with LEAP have consistent improvements across both CV and Graph Learning domains. We obtain 4.64 \% to 12.89\% relative error rate reduction in CV domain, and 6.10 \% to 15.26\% in Graph domain. 
Last, optimizers does not affect the effectiveness of LEAP. SGD with momentum and Adam are top-2 widely-used optimizer, and Table~\ref{tab:Optimizer} show that LEAP can achieve similar great results with both optimizers, which shows the significant practical potential of LEAP.

\section{Conclusion}
We propose a simple and effective learning rate schedule plugin, called \textbf{LEA}rning Rate \textbf{P}erturbation (LEAP).
Similar to Dropout and Batch Normalization, LEAP is a generic plugin for improving the performance of deep learning models.
We found that LEAP make optimizers prefer flatter minima to sharp minima.
Experiments show that training with LEAP improve the performance of various deep learning models on diverse datasets of different domains.

\section{Acknowledgments}
Hengyu Liu’s work described in this paper was in part supported by the computing resources funded by National Natural Science Foundation of China (Nos.U1811261, 62137001). 
\bibliographystyle{ACM-Reference-Format}
\balance
\bibliography{main}


\begin{thebibliography}{39}

%%% the Web address.

\ifx \showCODEN    \undefined \def \showCODEN     #1{\unskip}     \fi
\ifx \showDOI      \undefined \def \showDOI       #1{#1}\fi
\ifx \showISBNx    \undefined \def \showISBNx     #1{\unskip}     \fi
\ifx \showISBNxiii \undefined \def \showISBNxiii  #1{\unskip}     \fi
\ifx \showISSN     \undefined \def \showISSN      #1{\unskip}     \fi
\ifx \showLCCN     \undefined \def \showLCCN      #1{\unskip}     \fi
\ifx \shownote     \undefined \def \shownote      #1{#1}          \fi
\ifx \showarticletitle \undefined \def \showarticletitle #1{#1}   \fi
\ifx \showURL      \undefined \def \showURL       {\relax}        \fi
% invisible to TeX
\providecommand\bibfield[2]{#2}
\providecommand\bibinfo[2]{#2}
\providecommand\natexlab[1]{#1}
\providecommand\showeprint[2][]{arXiv:#2}

\bibitem[\protect\citeauthoryear{Bishop}{Bishop}{1995}]%
        {bishop1995training}
\bibfield{author}{\bibinfo{person}{Chris~M Bishop}.}
  \bibinfo{year}{1995}\natexlab{}.
\newblock \showarticletitle{Training with noise is equivalent to Tikhonov
  regularization}.
\newblock \bibinfo{journal}{\emph{Neural computation}} \bibinfo{volume}{7},
  \bibinfo{number}{1} (\bibinfo{year}{1995}), \bibinfo{pages}{108--116}.
\newblock


\bibitem[\protect\citeauthoryear{Bishop et~al\mbox{.}}{Bishop
  et~al\mbox{.}}{1995}]%
        {bishop1995neural}
\bibfield{author}{\bibinfo{person}{Christopher~M Bishop} {et~al\mbox{.}}}
  \bibinfo{year}{1995}\natexlab{}.
\newblock \bibinfo{booktitle}{\emph{Neural networks for pattern recognition}}.
\newblock \bibinfo{publisher}{Oxford university press}.
\newblock


\bibitem[\protect\citeauthoryear{Chen, Du, Chen, Wang, Long, and Xie}{Chen
  et~al\mbox{.}}{2021}]%
        {chen2021fast}
\bibfield{author}{\bibinfo{person}{Xu Chen}, \bibinfo{person}{Lun Du},
  \bibinfo{person}{Mengyuan Chen}, \bibinfo{person}{Yun Wang},
  \bibinfo{person}{Qingqing Long}, {and} \bibinfo{person}{Kunqing Xie}.}
  \bibinfo{year}{2021}\natexlab{}.
\newblock \showarticletitle{Fast Hierarchy Preserving Graph Embedding via
  Subspace Constraints}. In \bibinfo{booktitle}{\emph{ICASSP 2021-2021 IEEE
  International Conference on Acoustics, Speech and Signal Processing
  (ICASSP)}}. IEEE, \bibinfo{pages}{3580--3584}.
\newblock


\bibitem[\protect\citeauthoryear{Deng, Dong, Socher, Li, Li, and Fei-Fei}{Deng
  et~al\mbox{.}}{2009}]%
        {deng2009imagenet}
\bibfield{author}{\bibinfo{person}{Jia Deng}, \bibinfo{person}{Wei Dong},
  \bibinfo{person}{Richard Socher}, \bibinfo{person}{Li-Jia Li},
  \bibinfo{person}{Kai Li}, {and} \bibinfo{person}{Li Fei-Fei}.}
  \bibinfo{year}{2009}\natexlab{}.
\newblock \showarticletitle{Imagenet: A large-scale hierarchical image
  database}. In \bibinfo{booktitle}{\emph{2009 IEEE conference on computer
  vision and pattern recognition}}. Ieee, \bibinfo{pages}{248--255}.
\newblock


\bibitem[\protect\citeauthoryear{Du, Chen, Gao, Fu, Xie, Han, and Zhang}{Du
  et~al\mbox{.}}{2022a}]%
        {du2022understanding}
\bibfield{author}{\bibinfo{person}{Lun Du}, \bibinfo{person}{Xu Chen},
  \bibinfo{person}{Fei Gao}, \bibinfo{person}{Qiang Fu},
  \bibinfo{person}{Kunqing Xie}, \bibinfo{person}{Shi Han}, {and}
  \bibinfo{person}{Dongmei Zhang}.} \bibinfo{year}{2022}\natexlab{a}.
\newblock \showarticletitle{Understanding and Improvement of Adversarial
  Training for Network Embedding from an Optimization Perspective}. In
  \bibinfo{booktitle}{\emph{Proceedings of the Fifteenth ACM International
  Conference on Web Search and Data Mining}}. \bibinfo{pages}{230--240}.
\newblock


\bibitem[\protect\citeauthoryear{Du, Gao, Chen, Jia, Wang, Zhang, Han, and
  Zhang}{Du et~al\mbox{.}}{2021}]%
        {du2021tabularnet}
\bibfield{author}{\bibinfo{person}{Lun Du}, \bibinfo{person}{Fei Gao},
  \bibinfo{person}{Xu Chen}, \bibinfo{person}{Ran Jia},
  \bibinfo{person}{Junshan Wang}, \bibinfo{person}{Jiang Zhang},
  \bibinfo{person}{Shi Han}, {and} \bibinfo{person}{Dongmei Zhang}.}
  \bibinfo{year}{2021}\natexlab{}.
\newblock \showarticletitle{TabularNet: A neural network architecture for
  understanding semantic structures of tabular data}. In
  \bibinfo{booktitle}{\emph{Proceedings of the 27th ACM SIGKDD Conference on
  Knowledge Discovery \& Data Mining}}. \bibinfo{pages}{322--331}.
\newblock


\bibitem[\protect\citeauthoryear{Du, Shi, Fu, Ma, Liu, Han, and Zhang}{Du
  et~al\mbox{.}}{2022b}]%
        {du2022gbk}
\bibfield{author}{\bibinfo{person}{Lun Du}, \bibinfo{person}{Xiaozhou Shi},
  \bibinfo{person}{Qiang Fu}, \bibinfo{person}{Xiaojun Ma},
  \bibinfo{person}{Hengyu Liu}, \bibinfo{person}{Shi Han}, {and}
  \bibinfo{person}{Dongmei Zhang}.} \bibinfo{year}{2022}\natexlab{b}.
\newblock \showarticletitle{GBK-GNN: Gated Bi-Kernel Graph Neural Networks for
  Modeling Both Homophily and Heterophily}. In
  \bibinfo{booktitle}{\emph{Proceedings of the ACM Web Conference 2022}}.
  \bibinfo{pages}{1550--1558}.
\newblock


\bibitem[\protect\citeauthoryear{Du, Song, Wang, Huang, Ruan, and Yu}{Du
  et~al\mbox{.}}{2018}]%
        {du2018traffic}
\bibfield{author}{\bibinfo{person}{Lun Du}, \bibinfo{person}{Guojie Song},
  \bibinfo{person}{Yiming Wang}, \bibinfo{person}{Jipeng Huang},
  \bibinfo{person}{Mengfei Ruan}, {and} \bibinfo{person}{Zhanyuan Yu}.}
  \bibinfo{year}{2018}\natexlab{}.
\newblock \showarticletitle{Traffic events oriented dynamic traffic assignment
  model for expressway network: a network flow approach}.
\newblock \bibinfo{journal}{\emph{IEEE Intelligent Transportation Systems
  Magazine}} \bibinfo{volume}{10}, \bibinfo{number}{1} (\bibinfo{year}{2018}),
  \bibinfo{pages}{107--120}.
\newblock


\bibitem[\protect\citeauthoryear{Fu, Du, Mao, Chen, Fang, Han, and Zhang}{Fu
  et~al\mbox{.}}{2021}]%
        {fu2021neuron}
\bibfield{author}{\bibinfo{person}{Qiang Fu}, \bibinfo{person}{Lun Du},
  \bibinfo{person}{Haitao Mao}, \bibinfo{person}{Xu Chen}, \bibinfo{person}{Wei
  Fang}, \bibinfo{person}{Shi Han}, {and} \bibinfo{person}{Dongmei Zhang}.}
  \bibinfo{year}{2021}\natexlab{}.
\newblock \showarticletitle{Neuron with Steady Response Leads to Better
  Generalization}.
\newblock \bibinfo{journal}{\emph{arXiv preprint arXiv:2111.15414}}
  (\bibinfo{year}{2021}).
\newblock


\bibitem[\protect\citeauthoryear{Gulcehre, Moczulski, Denil, and
  Bengio}{Gulcehre et~al\mbox{.}}{2016}]%
        {gulcehre2016noisy}
\bibfield{author}{\bibinfo{person}{Caglar Gulcehre}, \bibinfo{person}{Marcin
  Moczulski}, \bibinfo{person}{Misha Denil}, {and} \bibinfo{person}{Yoshua
  Bengio}.} \bibinfo{year}{2016}\natexlab{}.
\newblock \showarticletitle{Noisy activation functions}. In
  \bibinfo{booktitle}{\emph{International conference on machine learning}}.
  PMLR, \bibinfo{pages}{3059--3068}.
\newblock


\bibitem[\protect\citeauthoryear{Hamilton, Ying, and Leskovec}{Hamilton
  et~al\mbox{.}}{2017}]%
        {DBLP:journals/corr/HamiltonYL17}
\bibfield{author}{\bibinfo{person}{William~L. Hamilton}, \bibinfo{person}{Rex
  Ying}, {and} \bibinfo{person}{Jure Leskovec}.}
  \bibinfo{year}{2017}\natexlab{}.
\newblock \showarticletitle{Inductive Representation Learning on Large Graphs}.
\newblock \bibinfo{journal}{\emph{CoRR}}  \bibinfo{volume}{abs/1706.02216}
  (\bibinfo{year}{2017}).
\newblock
\showeprint[arXiv]{1706.02216}
\urldef\tempurl%
\url{http://arxiv.org/abs/1706.02216}
\showURL{%
\tempurl}


\bibitem[\protect\citeauthoryear{Haykin}{Haykin}{1994}]%
        {haykin1994neural}
\bibfield{author}{\bibinfo{person}{Simon Haykin}.}
  \bibinfo{year}{1994}\natexlab{}.
\newblock \bibinfo{booktitle}{\emph{Neural networks: a comprehensive
  foundation}}.
\newblock \bibinfo{publisher}{Prentice Hall PTR}.
\newblock


\bibitem[\protect\citeauthoryear{He, Zhang, Ren, and Sun}{He
  et~al\mbox{.}}{2016}]%
        {he2016deep}
\bibfield{author}{\bibinfo{person}{Kaiming He}, \bibinfo{person}{Xiangyu
  Zhang}, \bibinfo{person}{Shaoqing Ren}, {and} \bibinfo{person}{Jian Sun}.}
  \bibinfo{year}{2016}\natexlab{}.
\newblock \showarticletitle{Deep residual learning for image recognition}. In
  \bibinfo{booktitle}{\emph{Proceedings of the IEEE conference on computer
  vision and pattern recognition}}. \bibinfo{pages}{770--778}.
\newblock


\bibitem[\protect\citeauthoryear{Kingma and Ba}{Kingma and Ba}{2014}]%
        {kingma2014adam}
\bibfield{author}{\bibinfo{person}{Diederik~P Kingma} {and}
  \bibinfo{person}{Jimmy Ba}.} \bibinfo{year}{2014}\natexlab{}.
\newblock \showarticletitle{Adam: A method for stochastic optimization}.
\newblock \bibinfo{journal}{\emph{arXiv preprint arXiv:1412.6980}}
  (\bibinfo{year}{2014}).
\newblock


\bibitem[\protect\citeauthoryear{Kipf and Welling}{Kipf and Welling}{2016}]%
        {kipf2016semi}
\bibfield{author}{\bibinfo{person}{Thomas~N Kipf} {and} \bibinfo{person}{Max
  Welling}.} \bibinfo{year}{2016}\natexlab{}.
\newblock \showarticletitle{Semi-supervised classification with graph
  convolutional networks}.
\newblock \bibinfo{journal}{\emph{arXiv preprint arXiv:1609.02907}}
  (\bibinfo{year}{2016}).
\newblock


\bibitem[\protect\citeauthoryear{Krizhevsky, Hinton, et~al\mbox{.}}{Krizhevsky
  et~al\mbox{.}}{2009}]%
        {krizhevsky2009learning}
\bibfield{author}{\bibinfo{person}{Alex Krizhevsky}, \bibinfo{person}{Geoffrey
  Hinton}, {et~al\mbox{.}}} \bibinfo{year}{2009}\natexlab{}.
\newblock \showarticletitle{Learning multiple layers of features from tiny
  images}.
\newblock  (\bibinfo{year}{2009}).
\newblock


\bibitem[\protect\citeauthoryear{LeCun}{LeCun}{1998}]%
        {lecun1998mnist}
\bibfield{author}{\bibinfo{person}{Yann LeCun}.}
  \bibinfo{year}{1998}\natexlab{}.
\newblock \showarticletitle{The MNIST database of handwritten digits}.
\newblock \bibinfo{journal}{\emph{http://yann. lecun. com/exdb/mnist/}}
  (\bibinfo{year}{1998}).
\newblock


\bibitem[\protect\citeauthoryear{Lewkowycz}{Lewkowycz}{2021}]%
        {lewkowycz2021decay}
\bibfield{author}{\bibinfo{person}{Aitor Lewkowycz}.}
  \bibinfo{year}{2021}\natexlab{}.
\newblock \showarticletitle{How to decay your learning rate}.
\newblock \bibinfo{journal}{\emph{arXiv preprint arXiv:2103.12682}}
  (\bibinfo{year}{2021}).
\newblock


\bibitem[\protect\citeauthoryear{Liu, Jiang, He, Chen, Liu, Gao, and Han}{Liu
  et~al\mbox{.}}{2019}]%
        {liu2019variance}
\bibfield{author}{\bibinfo{person}{Liyuan Liu}, \bibinfo{person}{Haoming
  Jiang}, \bibinfo{person}{Pengcheng He}, \bibinfo{person}{Weizhu Chen},
  \bibinfo{person}{Xiaodong Liu}, \bibinfo{person}{Jianfeng Gao}, {and}
  \bibinfo{person}{Jiawei Han}.} \bibinfo{year}{2019}\natexlab{}.
\newblock \showarticletitle{On the variance of the adaptive learning rate and
  beyond}.
\newblock \bibinfo{journal}{\emph{arXiv preprint arXiv:1908.03265}}
  (\bibinfo{year}{2019}).
\newblock


\bibitem[\protect\citeauthoryear{Liu, Lin, Cao, Hu, Wei, Zhang, Lin, and
  Guo}{Liu et~al\mbox{.}}{2021}]%
        {liu2021swin}
\bibfield{author}{\bibinfo{person}{Ze Liu}, \bibinfo{person}{Yutong Lin},
  \bibinfo{person}{Yue Cao}, \bibinfo{person}{Han Hu}, \bibinfo{person}{Yixuan
  Wei}, \bibinfo{person}{Zheng Zhang}, \bibinfo{person}{Stephen Lin}, {and}
  \bibinfo{person}{Baining Guo}.} \bibinfo{year}{2021}\natexlab{}.
\newblock \showarticletitle{Swin transformer: Hierarchical vision transformer
  using shifted windows}.
\newblock \bibinfo{journal}{\emph{arXiv preprint arXiv:2103.14030}}
  (\bibinfo{year}{2021}).
\newblock


\bibitem[\protect\citeauthoryear{Loshchilov and Hutter}{Loshchilov and
  Hutter}{2016}]%
        {loshchilov2016sgdr}
\bibfield{author}{\bibinfo{person}{Ilya Loshchilov} {and}
  \bibinfo{person}{Frank Hutter}.} \bibinfo{year}{2016}\natexlab{}.
\newblock \showarticletitle{Sgdr: Stochastic gradient descent with warm
  restarts}.
\newblock \bibinfo{journal}{\emph{arXiv preprint arXiv:1608.03983}}
  (\bibinfo{year}{2016}).
\newblock


\bibitem[\protect\citeauthoryear{Mao, Chen, Fu, Du, Han, and Zhang}{Mao
  et~al\mbox{.}}{2021}]%
        {mao2021neuron}
\bibfield{author}{\bibinfo{person}{Haitao Mao}, \bibinfo{person}{Xu Chen},
  \bibinfo{person}{Qiang Fu}, \bibinfo{person}{Lun Du}, \bibinfo{person}{Shi
  Han}, {and} \bibinfo{person}{Dongmei Zhang}.}
  \bibinfo{year}{2021}\natexlab{}.
\newblock \showarticletitle{Neuron Campaign for Initialization Guided by
  Information Bottleneck Theory}. In \bibinfo{booktitle}{\emph{Proceedings of
  the 30th ACM International Conference on Information \& Knowledge
  Management}}. \bibinfo{pages}{3328--3332}.
\newblock


\bibitem[\protect\citeauthoryear{Mishra and Sarawadekar}{Mishra and
  Sarawadekar}{2019}]%
        {mishra2019polynomial}
\bibfield{author}{\bibinfo{person}{Purnendu Mishra} {and}
  \bibinfo{person}{Kishor Sarawadekar}.} \bibinfo{year}{2019}\natexlab{}.
\newblock \showarticletitle{Polynomial learning rate policy with warm restart
  for deep neural network}. In \bibinfo{booktitle}{\emph{TENCON 2019-2019 IEEE
  Region 10 Conference (TENCON)}}. IEEE, \bibinfo{pages}{2087--2092}.
\newblock


\bibitem[\protect\citeauthoryear{Neelakantan, Vilnis, Le, Sutskever, Kaiser,
  Kurach, and Martens}{Neelakantan et~al\mbox{.}}{2015}]%
        {neelakantan2015adding}
\bibfield{author}{\bibinfo{person}{Arvind Neelakantan}, \bibinfo{person}{Luke
  Vilnis}, \bibinfo{person}{Quoc~V Le}, \bibinfo{person}{Ilya Sutskever},
  \bibinfo{person}{Lukasz Kaiser}, \bibinfo{person}{Karol Kurach}, {and}
  \bibinfo{person}{James Martens}.} \bibinfo{year}{2015}\natexlab{}.
\newblock \showarticletitle{Adding gradient noise improves learning for very
  deep networks}.
\newblock \bibinfo{journal}{\emph{arXiv preprint arXiv:1511.06807}}
  (\bibinfo{year}{2015}).
\newblock


\bibitem[\protect\citeauthoryear{Preechakul and Kijsirikul}{Preechakul and
  Kijsirikul}{2019}]%
        {preechakul2019cprop}
\bibfield{author}{\bibinfo{person}{Konpat Preechakul} {and}
  \bibinfo{person}{Boonserm Kijsirikul}.} \bibinfo{year}{2019}\natexlab{}.
\newblock \showarticletitle{CProp: Adaptive Learning Rate Scaling from Past
  Gradient Conformity}.
\newblock \bibinfo{journal}{\emph{arXiv preprint arXiv:1912.11493}}
  (\bibinfo{year}{2019}).
\newblock


\bibitem[\protect\citeauthoryear{Reed and MarksII}{Reed and MarksII}{1999}]%
        {reed1999neural}
\bibfield{author}{\bibinfo{person}{Russell Reed} {and}
  \bibinfo{person}{Robert~J MarksII}.} \bibinfo{year}{1999}\natexlab{}.
\newblock \bibinfo{booktitle}{\emph{Neural smithing: supervised learning in
  feedforward artificial neural networks}}.
\newblock \bibinfo{publisher}{Mit Press}.
\newblock


\bibitem[\protect\citeauthoryear{Ruder}{Ruder}{2016}]%
        {ruder2016overview}
\bibfield{author}{\bibinfo{person}{Sebastian Ruder}.}
  \bibinfo{year}{2016}\natexlab{}.
\newblock \showarticletitle{An overview of gradient descent optimization
  algorithms}.
\newblock \bibinfo{journal}{\emph{arXiv preprint arXiv:1609.04747}}
  (\bibinfo{year}{2016}).
\newblock


\bibitem[\protect\citeauthoryear{Russakovsky, Deng, Su, Krause, Satheesh, Ma,
  Huang, Karpathy, Khosla, Bernstein, et~al\mbox{.}}{Russakovsky
  et~al\mbox{.}}{2015}]%
        {russakovsky2015imagenet}
\bibfield{author}{\bibinfo{person}{Olga Russakovsky}, \bibinfo{person}{Jia
  Deng}, \bibinfo{person}{Hao Su}, \bibinfo{person}{Jonathan Krause},
  \bibinfo{person}{Sanjeev Satheesh}, \bibinfo{person}{Sean Ma},
  \bibinfo{person}{Zhiheng Huang}, \bibinfo{person}{Andrej Karpathy},
  \bibinfo{person}{Aditya Khosla}, \bibinfo{person}{Michael Bernstein},
  {et~al\mbox{.}}} \bibinfo{year}{2015}\natexlab{}.
\newblock \showarticletitle{Imagenet large scale visual recognition challenge}.
\newblock \bibinfo{journal}{\emph{International journal of computer vision}}
  \bibinfo{volume}{115}, \bibinfo{number}{3} (\bibinfo{year}{2015}),
  \bibinfo{pages}{211--252}.
\newblock


\bibitem[\protect\citeauthoryear{Shi, Wang, Du, Chen, Han, Zhang, Zhang, and
  Sun}{Shi et~al\mbox{.}}{2021a}]%
        {shi2021neural}
\bibfield{author}{\bibinfo{person}{Ensheng Shi}, \bibinfo{person}{Yanlin Wang},
  \bibinfo{person}{Lun Du}, \bibinfo{person}{Junjie Chen}, \bibinfo{person}{Shi
  Han}, \bibinfo{person}{Hongyu Zhang}, \bibinfo{person}{Dongmei Zhang}, {and}
  \bibinfo{person}{Hongbin Sun}.} \bibinfo{year}{2021}\natexlab{a}.
\newblock \showarticletitle{Neural Code Summarization: How Far Are We?}
\newblock \bibinfo{journal}{\emph{arXiv preprint arXiv:2107.07112}}
  (\bibinfo{year}{2021}).
\newblock


\bibitem[\protect\citeauthoryear{Shi, Wang, Du, Zhang, Han, Zhang, and Sun}{Shi
  et~al\mbox{.}}{2021b}]%
        {shi2021cast}
\bibfield{author}{\bibinfo{person}{Ensheng Shi}, \bibinfo{person}{Yanlin Wang},
  \bibinfo{person}{Lun Du}, \bibinfo{person}{Hongyu Zhang},
  \bibinfo{person}{Shi Han}, \bibinfo{person}{Dongmei Zhang}, {and}
  \bibinfo{person}{Hongbin Sun}.} \bibinfo{year}{2021}\natexlab{b}.
\newblock \showarticletitle{Cast: Enhancing code summarization with
  hierarchical splitting and reconstruction of abstract syntax trees}.
\newblock \bibinfo{journal}{\emph{arXiv preprint arXiv:2108.12987}}
  (\bibinfo{year}{2021}).
\newblock


\bibitem[\protect\citeauthoryear{Simonyan and Zisserman}{Simonyan and
  Zisserman}{2014}]%
        {simonyan2014very}
\bibfield{author}{\bibinfo{person}{Karen Simonyan} {and}
  \bibinfo{person}{Andrew Zisserman}.} \bibinfo{year}{2014}\natexlab{}.
\newblock \showarticletitle{Very deep convolutional networks for large-scale
  image recognition}.
\newblock \bibinfo{journal}{\emph{arXiv preprint arXiv:1409.1556}}
  (\bibinfo{year}{2014}).
\newblock


\bibitem[\protect\citeauthoryear{Smith}{Smith}{2017}]%
        {smith2017cyclical}
\bibfield{author}{\bibinfo{person}{Leslie~N Smith}.}
  \bibinfo{year}{2017}\natexlab{}.
\newblock \showarticletitle{Cyclical learning rates for training neural
  networks}. In \bibinfo{booktitle}{\emph{2017 IEEE winter conference on
  applications of computer vision (WACV)}}. IEEE, \bibinfo{pages}{464--472}.
\newblock


\bibitem[\protect\citeauthoryear{Song, Li, Wang, and Du}{Song
  et~al\mbox{.}}{2020}]%
        {song2020inferring}
\bibfield{author}{\bibinfo{person}{Guojie Song}, \bibinfo{person}{Yuanhao Li},
  \bibinfo{person}{Junshan Wang}, {and} \bibinfo{person}{Lun Du}.}
  \bibinfo{year}{2020}\natexlab{}.
\newblock \showarticletitle{Inferring explicit and implicit social ties
  simultaneously in mobile social networks}.
\newblock \bibinfo{journal}{\emph{Science China Information Sciences}}
  \bibinfo{volume}{63}, \bibinfo{number}{4} (\bibinfo{year}{2020}),
  \bibinfo{pages}{1--3}.
\newblock


\bibitem[\protect\citeauthoryear{Veli{\v{c}}kovi{\'c}, Cucurull, Casanova,
  Romero, Lio, and Bengio}{Veli{\v{c}}kovi{\'c} et~al\mbox{.}}{2017}]%
        {velivckovic2017graph}
\bibfield{author}{\bibinfo{person}{Petar Veli{\v{c}}kovi{\'c}},
  \bibinfo{person}{Guillem Cucurull}, \bibinfo{person}{Arantxa Casanova},
  \bibinfo{person}{Adriana Romero}, \bibinfo{person}{Pietro Lio}, {and}
  \bibinfo{person}{Yoshua Bengio}.} \bibinfo{year}{2017}\natexlab{}.
\newblock \showarticletitle{Graph attention networks}.
\newblock \bibinfo{journal}{\emph{arXiv preprint arXiv:1710.10903}}
  (\bibinfo{year}{2017}).
\newblock


\bibitem[\protect\citeauthoryear{Xie, Sato, and Sugiyama}{Xie
  et~al\mbox{.}}{2020}]%
        {xie2020diffusion}
\bibfield{author}{\bibinfo{person}{Zeke Xie}, \bibinfo{person}{Issei Sato},
  {and} \bibinfo{person}{Masashi Sugiyama}.} \bibinfo{year}{2020}\natexlab{}.
\newblock \showarticletitle{A diffusion theory for deep learning dynamics:
  Stochastic gradient descent exponentially favors flat minima}.
\newblock \bibinfo{journal}{\emph{arXiv preprint arXiv:2002.03495}}
  (\bibinfo{year}{2020}).
\newblock


\bibitem[\protect\citeauthoryear{Xu, Hu, Leskovec, and Jegelka}{Xu
  et~al\mbox{.}}{2018}]%
        {xu2018powerful}
\bibfield{author}{\bibinfo{person}{Keyulu Xu}, \bibinfo{person}{Weihua Hu},
  \bibinfo{person}{Jure Leskovec}, {and} \bibinfo{person}{Stefanie Jegelka}.}
  \bibinfo{year}{2018}\natexlab{}.
\newblock \showarticletitle{How powerful are graph neural networks?}
\newblock \bibinfo{journal}{\emph{arXiv preprint arXiv:1810.00826}}
  (\bibinfo{year}{2018}).
\newblock


\bibitem[\protect\citeauthoryear{Yang, Cohen, and Salakhudinov}{Yang
  et~al\mbox{.}}{2016}]%
        {yang2016revisiting}
\bibfield{author}{\bibinfo{person}{Zhilin Yang}, \bibinfo{person}{William
  Cohen}, {and} \bibinfo{person}{Ruslan Salakhudinov}.}
  \bibinfo{year}{2016}\natexlab{}.
\newblock \showarticletitle{Revisiting semi-supervised learning with graph
  embeddings}. In \bibinfo{booktitle}{\emph{International conference on machine
  learning}}. PMLR, \bibinfo{pages}{40--48}.
\newblock


\bibitem[\protect\citeauthoryear{You, Long, Wang, and Jordan}{You
  et~al\mbox{.}}{2019}]%
        {you2019does}
\bibfield{author}{\bibinfo{person}{Kaichao You}, \bibinfo{person}{Mingsheng
  Long}, \bibinfo{person}{Jianmin Wang}, {and} \bibinfo{person}{Michael~I
  Jordan}.} \bibinfo{year}{2019}\natexlab{}.
\newblock \showarticletitle{How does learning rate decay help modern neural
  networks?}
\newblock \bibinfo{journal}{\emph{arXiv preprint arXiv:1908.01878}}
  (\bibinfo{year}{2019}).
\newblock


\bibitem[\protect\citeauthoryear{Zeiler}{Zeiler}{2012}]%
        {zeiler2012adadelta}
\bibfield{author}{\bibinfo{person}{Matthew~D Zeiler}.}
  \bibinfo{year}{2012}\natexlab{}.
\newblock \showarticletitle{Adadelta: an adaptive learning rate method}.
\newblock \bibinfo{journal}{\emph{arXiv preprint arXiv:1212.5701}}
  (\bibinfo{year}{2012}).
\newblock


\end{thebibliography}
\appendix
\section{Details of Reproduction} 
\label{app:reproduction}
% \subsection{Open Source Code}
% We publish our code in an anonymous code repository (i.e., \url{https://anonymous.4open.science/r/}).
\subsection{Datasets} \label{app:datasets}
We take 6 real datasets from Computer Vision and Graph Learning domain to evaluate our method. The detailed descriptions are listed as follows:
\begin{itemize}
    \item \textbf{Cora, CiteSeer and PubMed} \cite{yang2016revisiting} are citation networks based datasets. In these datasets, nodes represent papers, and edges represent citations of one paper by another. Node features are the bag-of-words representation of papers, and node label denotes the academic topic of a paper.
    \item \textbf{MNIST} \cite{lecun1998mnist} MNIST is one of the most researched datasets in machine learning, and is used to classify handwritten digits. 
    \item \textbf{CIFAR-10} \cite{krizhevsky2009learning} The CIFAR-10 dataset is a collection of images that are commonly used to train machine learning and computer vision algorithms. It is one of the most widely used datasets for machine learning research. The CIFAR-10 dataset contains 60,000 32x32 color images in 10 different classes. The 10 different classes represent airplanes, cars, birds, etc.
    \item \textbf{ImageNet} \cite{deng2009imagenet} The ImageNet is a large visual database designed for use in visual object recognition software research. More than 14 million images have been hand-annotated by the project to indicate what objects are pictured and in at least one million of the images, bounding boxes are also provided. ImageNet contains more than 20,000 categories with a typical category, such as "balloon" or "strawberry", consisting of several hundred images.
\end{itemize}
 % 数据集划分
    % GNN
        % Cora, CiteSeer and PubMed 与原始论文相同
For graph datasets, we use the public data splits provided by \cite{yang2016revisiting}. 
    % CV
        % MNIST [50000 10000 10000]
        % Cifar-10 
        % ImageNet
For the MNIST dataset, we keep the test set unchanged, and we randomly select 50,000 out of 60,000 training images for training and the other 10,000 images as validation set for hyper-parameter tuning.
For Cifar-10 and ImageNet dataset, we use the public data splits provided by \cite{krizhevsky2009learning,russakovsky2015imagenet}.
\subsection{Network Architechtures} \label{app:models}
To verify the effectiveness of our method across different neural architectures, we employ Multilayer Perceptrons (MLPs), Convolutional Neural Networks (CNNs), Transforms, and Graph Neural Networks (GNNs) for evaluation. 

\begin{itemize}
    \item \textbf{MLP} \cite{haykin1994neural} is the most classic neural network with multiple layers between the input and output layers. 
    % There are different types of neural networks but they always consist of the same components: neurons, synapses, weights, biases, and functions. These components functioning similar to the human brains.
    \item \textbf{ResNet} \cite{he2016deep} is a neural network of a kind that builds on constructs known from pyramidal cells in the cerebral cortex. ResNet do this by utilizing skip connections, or shortcuts to jump over some layers.
    \item \textbf{VGG} \cite{simonyan2014very} is a Convolutional Neural Network architcture, it was submitted to Large Scale Visual Recognition Challenge 2014 (ILSVRC2014). 
    \item \textbf{Swin Transfermer} \cite{liu2021swin} 
    % capably serves as a general-purpose backbone for computer vision. It 
    is basically a hierarchical Transformer whose representation is computed with shifted windows. 
    % The shifted windowing scheme brings greater efficiency by limiting self-attention computation to non-overlapping local windows while also allowing for cross-window connection.
    \item \textbf{GCN} \cite{kipf2016semi} is a semi-supervised graph convolutional network model which learns node representations by aggregating information from neighbors.
    \item \textbf{GAT} \cite{velivckovic2017graph} is a graph neural network model using attention mechanism to aggregate node features.
    \item \textbf{GIN} \cite{xu2018powerful} is a graph neural network which can distinguish graph structures more as powerful as the Weisfeiler-Lehman test.
    \item \textbf{GraphSage} \cite{DBLP:journals/corr/HamiltonYL17} is a general inductive framework that leverages node feature information to efficiently generate node embedding for previously unseen data.
\end{itemize}

For MLPs, we run five MLP models, which are denoted as MLP-L, L indicates the number of layers including input layer. The detailed hidden size is listed in the Appendix. We adopt ResNet-18, ResNet-50, ResNet-101 \cite{he2016deep} and VGG-19 \cite{simonyan2014very} for CNNs, Swin Transformer \cite{liu2021swin} for Transformer, and GCN \cite{kipf2016semi}, GAT \cite{velivckovic2017graph}, GIN \cite{xu2018powerful}, and GraphSage \cite{DBLP:journals/corr/HamiltonYL17} for GNNs. ResNet and VGG are two conventional CNN models, Swin is a Transformer-based SOTA model for CV-related tasks, and GCN, GAT, GIN, and GraphSage are four widely used GNN models for the node classification task. To be specific, we use 2-layer GNN models with hidden size 16 and Tiny Swin Transfermor \cite{liu2021swin}.

\begin{table}[!ht]
    \centering
    \begin{tabular}{c|c}
     \hline
     Model    &    Hidden layer dimension              \\
     \hline
     MLP-3    &    [784, 100, 10] \\
     MLP-4    &    [784, 256, 100, 10]                 \\
     MLP-6    &    [784, 256, 128, 64, 32, 10]         \\
     MLP-8    &    [784, 256, 128, 64, 64, 32, 32, 10] \\
     MLP-10   &    [784, 256, 128, 64, 64, 32, 32, 16, 16, 10] \\
     \hline
    \end{tabular}
    \caption{The detail architectures of MLP}
    \label{tab:MLPARCH}
\end{table}

% We verify that LEAP improves the performance of six CV and GNN models. The six CV and GNN models are:
% % This experiment verifies that LEAP improves the performance of MLP, ResNet, VGG and Swin Transfermer models in CV domain, and verifies the performance improvement of LEAP for GCN, GAT, GIN and GraphSage models in GNN domain.
\subsection{Learning Rate Schedule} \label{app:lrs}
We verify that LEAP improves the performance of 6 CV and GNN models on three different learning rate schedules. The three different learning rate schedules are:
\begin{itemize}
    \item \textbf{Constant Learning Rate} uses the same learning rate during the whole training process. Actually, it can be considered as training without learning rate schedule. 
    % \lun{Actually, it's not a schedule. It means we don't use any schedule. I think we can remove the term.}
    \item \textbf{Learning Rate Decay} selects an initial learning rate, then gradually reduce it in accordance with a schedule.
    % \footnote{\url{https://pytorch.org/docs/stable/generated/torch.optim.lr_scheduler.MultiStepLR.html#torch.optim.lr_scheduler.MultiStepLR}} 
    % \item \textbf{Cyclical Schedule} \cite{smith2017cyclical} lets the learning rate cyclically vary between reasonable boundary values.
    \item \textbf{Warm Restart Schedule} \cite{loshchilov2016sgdr} is a simple warm restart technique for stochastic gradient descent to improve its anytime performance when training deep neural networks.
\end{itemize}
\subsection{Optimizer} \label{app:optimizers}
Since our method LEAP directly interferes with the optimization process, how the optimizer affects the proposed method becomes an important research problem. We adopt Stochastic Gradient Descent (SGD) \cite{ruder2016overview} with momentum (hyperparameter $\beta = 0.9$) and its widely-used variant Adam \cite{kingma2014adam} to test the effects of different optimizers.
\subsection{Hyperparameter Settings.}
\label{app:hyperparameter}
% Swin-Tranformer and VGG-19 使用了现有参数
We use the hyperparameter of swin-transformer and VGG-19 provided by Model Zoo \footnote{\url{https://mmclassification.readthedocs.io/en/latest/model_zoo.html}}.
For the models without classical hyperparameter settings, we performed a hyperparameter search to find the best one for the baseline model. And the hyperparameter search process ensure each model in the same domain has the same search space. 
% GNN
    % learning rate: [0.1, 0.05, 0.01, 0.005, 0.001]
    % weight decay: [0.1, 0.05, 0.01, 0.005, 0.001, 0.0005, 0.0001]
For all the GNN models, learning rate is searched in $\{0.1, 0.05, 0.01, 5e-3, 1e-3\}$, and weight decay is searched in $\{0.1, 0.05, 0.01, 5e-3, 1e-3, 5e-4, 1e-4\}$.
% ResNet 
    % learning rate: [0.1, 0.05, 0.01]
    % weight decay: [0.005, 0.001, 0.0005, 0.0001]
For ResNet and MLP, learning rate is searched in $\{0.1, 0.05, 0.01\}$, and weight decay is searched in $\{5e-3, 1e-3, 5e-4, 1e-4\}$.
% \lamda [0.1 0.05 0.01 5e-3 1e-3 5e-4 1e-4 5e-5 1e-5]
Above is the hyperparameter setting and hyperparameter search for baselines. For LEAP, we search for the hyperparameter $\sigma$ in $\{0.1, 0.05, 0.01, 5e-3, 1e-3, 5e-4, 1e-4\}$ while keeping other hyperparameters unchanged. 
Note that we did not apply constant learning rate and learning rate decay for Swim Transformer because that these two learning rate schedules are not suitable for Swim Transformer. In fact, the performance of Swin Transformer which use these two learning rate schedules degrades greatly. 
We run each experiment on five random seeds and make an average. Model training is done on Nvidia Tesla V100 GPU.

\end{document}